\documentclass[10pt,twocolumn,letterpaper]{article}

\usepackage{confsty}
\usepackage{graphicx}
\usepackage{amsmath}
\usepackage{amssymb}
\usepackage{algorithm}
\usepackage{algpseudocode}
\usepackage{booktabs}
\usepackage{makecell}
\usepackage[pagebackref,breaklinks,colorlinks]{hyperref}
\usepackage[capitalize]{cleveref}
\crefname{section}{Sec.}{Secs.}
\Crefname{section}{Section}{Sections}
\Crefname{table}{Table}{Tables}
\crefname{table}{Tab.}{Tabs.}

\begin{document}

\title{evMLP: An Efficient Event-Driven MLP Architecture for Vision}

\author{
  Zhentan~Zheng\thanks{Corresponding author: xjtu\_evi@stu.xjtu.edu.cn.}\\
  \\
  Institute of Artificial Intelligence and Robotics, Xi’an Jiaotong University
}

\maketitle

\begin{abstract}

While CNNs and ViTs dominate vision architectures, all-MLP models offer a structurally simpler alternative whose patch-independent processing is naturally suited to exploiting temporal redundancy in video.
We present evMLP, an all-MLP architecture that processes image patches independently, enabling an event-driven local update mechanism for video processing: by defining inter-frame changes as ``events'' and processing only the patches where events occur, evMLP avoids redundant computation on unchanged regions.
Because each patch is processed independently, skipping an unchanged patch leaves all other outputs unaffected; at an event threshold of zero, the mechanism produces outputs identical to the dense baseline rather than an approximation.
On ImageNet, evMLP achieves 73.5\% top-1 accuracy at 1.03 GMACs (rising to 77.0\% with knowledge distillation and an extended training schedule). On multiple video datasets, the event-driven mechanism reduces computational cost by 8.4\%--26.8\% while maintaining output consistency with the dense baseline.
Wall-clock measurements confirm that these savings translate into actual speedup under compute-bound conditions, and that stream-level parallelism is the effective deployment strategy for multi-core systems.
The code and pre-trained models are available at \url{https://github.com/i-evi/evMLP}.
\end{abstract}

\section{Introduction}
\label{sec::intro}

Convolutional Neural Networks(CNNs) have achieved remarkable success in vision tasks\cite{ResNet,EfficientNet,ResNext}.
By scanning convolutional kernels across entire images, they effectively capture local features while reducing parameters through weight sharing.
Over more than a decade of development, CNNs have driven transformative progress across computer vision applications.
Recently, Vision Transformers (ViTs)\cite{ViT} have surpassed CNNs in multiple computer vision tasks by leveraging global modeling capabilities from self-attention mechanisms, emerging as a major research focus.
Additionally, novel architectures employing solely multi-layer perceptrons (MLPs) for image processing via fully-connected layers\cite{MLPMixer,ResMLP} offer fresh perspectives for the research on vision model architectures.

\begin{figure}[ht]
    \centering
    \begin{minipage}[t]{0.24\linewidth}
    \includegraphics[width=1.0\textwidth]{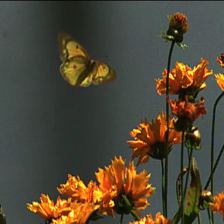}
    \centerline{\footnotesize (a)}
    \end{minipage}
    \begin{minipage}[t]{0.24\linewidth}
    \includegraphics[width=1.0\textwidth]{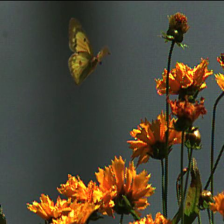}
    \centerline{\footnotesize (b)}
    \end{minipage}
    \begin{minipage}[t]{0.24\linewidth}
    \includegraphics[width=1.0\textwidth]{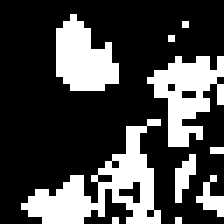}
    \centerline{\footnotesize (c)}
    \end{minipage}
    \begin{minipage}[t]{0.24\linewidth}
    \includegraphics[width=1.0\textwidth]{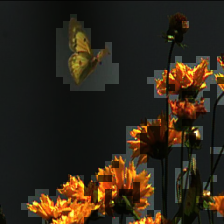}
    \centerline{\footnotesize (d)}
    \end{minipage}
    \caption{
        The event-driven local update mechanism.
        (a) and (b) are two consecutive frames with a resolution of $224\!\times\!224$.
        (c) presents the corresponding event map for a patch size of 7, where white regions denote activated events (the event calculation is detailed in Section \ref{sec::method::evupdate}).
        (d) is the current frame (b) masked by the event map (c), highlighting regions requiring recomputation.}
    \label{fig::intro}
\end{figure}

When vision models process image sequences such as videos, consecutive frames often contain redundant information from unchanged regions.
In this paper, we define changed areas between consecutive frames as \emph{events} and focus on patches where events occur.
This definition is inspired by the same principle underlying event cameras~\cite{EventCamera}, which asynchronously report per-pixel brightness changes; our work can be viewed as applying this change-driven paradigm to conventional frame-based input at the patch level.
We propose a MLP-based network architecture named evMLP and design a simple \emph{event-driven local update mechanism}, illustrated in Fig.~\ref{fig::intro}.
An \emph{event threshold} is introduced to determine event occurrence on corresponding patches between consecutive frames.
When no event occurs in a patch relative to its corresponding patch in the previous frame, our evMLP can reuse computation results from the prior frame's corresponding patch.
This avoids redundant calculations, thus improving network efficiency.

The key to implementing our proposed event-driven local update mechanism lies in the model's ability to process patches on images or feature maps independently.
In traditional ViTs or CNNs, processing patches on the image or feature map independently is non-trivial.
The attention mechanism is used to model global information in most ViTs, while the convolution kernels of CNNs scan across the entire image or feature map.
Therefore, we utilize MLPs to implement our network architecture.
Experimental results demonstrate that our evMLP achieves performance on par with mainstream state-of-the-art network architectures.
When trained from scratch on ImageNet-1K without extra data, it achieves a top-1 accuracy of 73.5\% at only 1.03 GMACs, outperforming similarly efficient models like DeiT-Ti\cite{DeiT} and T2T-ViT-7\cite{T2T}, rising to 77.0\% with knowledge distillation and an extended training schedule.
Through video processing on multiple datasets, we validated our event-driven local update mechanism.
We further conduct experiments to balance computational performance and output consistency via event threshold tuning.
Experimental results across multiple video datasets demonstrate that this mechanism enables our evMLP to achieve average computational cost reductions of 8.4\%--14.5\% in general scenarios without compromising output consistency.
Notably, in videos captured by stationary cameras, such as surveillance videos, it attains computational cost reductions exceeding 26\%.
By appropriately increasing the event threshold, the evMLP achieves greater computational efficiency gains while maintaining acceptable output consistency.
We further validate through wall-clock measurements that these computational savings translate into actual speedup, and identify stream-level parallelism as the effective deployment strategy for multi-core systems.

\section{Related Work}
\label{sec::related}

Since AlexNet\cite{AlexNet} achieved revolutionary results in the ImageNet challenge, CNNs have become the mainstream architecture.
The convolutional operation can efficiently extract local features of images, such as edges and textures.
VGG\cite{VGG} constructs deep features by stacking $3\!\times\!3$ convolutional kernels, and ResNet\cite{ResNet} breaks through the network degradation problem with residual connections.
Deeper stacks, larger parameter counts, and refined network topologies~\mbox{\cite{ResNext,DenseNet,EfficientNet}} have since driven CNN backbones to performance breakthroughs across a wide range of computer vision tasks.

In recent years, ViTs have achieved powerful global context modeling capabilities based on the self-attention mechanism\cite{Attention, ViT}.
Employing multi-head attention, ViTs capture both local and global dependencies within images, effectively overcoming the limitations of CNNs' local receptive fields.
DeiT\cite{DeiT} addresses ViTs' data-hungry nature via distillation, while Swin Transformer \cite{SwinViT} integrates global modeling capabilities with local processing efficiency through a shifted window approach.
Architectural innovations in Transformer-based vision models continue to drive performance breakthroughs, surpassing traditional CNNs across diverse computer vision tasks.

Furthermore, recent all-MLP architectures have achieved competitive performance using exclusively fully-connected layers\cite{MLPMixer,ResMLP}, offering novel perspectives for vision model design. Due to their structural simplicity, MLP-based models frequently attain higher inference throughput while matching the accuracy of conventional counterparts.
Subsequently, numerous MLP-like approaches have demonstrated promising results\cite{gMLP,ViP,StripMLP}, underscoring MLPs' significant potential as foundational building blocks for vision models.

Despite their impressive performance on visual tasks, deep neural networks incur substantial computational costs.
Static compression techniques like pruning\cite{Pruning,ThiNet}, quantization\cite{LimNumPrecision,AdaptiveQ}, and knowledge distillation\cite{Distilling,VarDistill} reduce model size while maintaining a fixed computational graph during inference.
In contrast, dynamic networks adapt their inference to the input to save resources, using strategies that depend on the architecture.
In cascaded models, computational depth is adjusted by skipping layers or exiting early; in parallel structures such as the Mixture of Experts (MoE)~\mbox{\cite{VMoE,AdaMV-MoE}}, a gating mechanism activates only a subset of experts per input, adjusting computational width.
Furthermore, methods like \cite{LookCloser,Glance} employ spatially-adaptive computation, selectively processing image areas to boost efficiency and performance.

Conventional algorithms have long utilized techniques such as motion compensation and optical flow to achieve efficient video coding and processing by exploiting inter-frame redundancy\cite{FlowEstim,H264}.
Subsequent deep learning approaches have further enhanced efficiency by modeling spatio-temporal information \cite{FlowNet,UniFormer}.
Nonetheless, these methods are frequently hampered by substantial computational expenses, such as those associated with optical flow estimation, or by intricate network architectures.
More recently, an emerging line of work has sought to directly exploit inter-frame redundancy within the model architecture itself.
Skip-Convolutions~\cite{SkipConv} restrict convolution to regions of significant change, DeltaCNN~\cite{DeltaCNN} propagates sparse frame-to-frame deltas through CNNs, CBInfer~\cite{CBInfer} thresholds feature changes for selective computation, and Eventful Transformers~\cite{EventfulTransformer} gate token updates in vision transformers according to temporal change.
These approaches share our goal of avoiding redundant computation on unchanged regions, yet differ from evMLP in a crucial aspect: because convolutions have spatially overlapping receptive fields and self-attention couples all tokens, skipping one location inevitably perturbs the outputs of others, so correctness must be maintained through approximate mechanisms such as delta accumulation or gating.
In contrast, evMLP processes each patch independently within a stage, so a skipped patch leaves the outputs of all other patches unaffected.
At $\tau = 0$, the event-driven mechanism thus produces outputs that are identical to those of its dense baseline, rather than an approximation of it. This property follows directly from the patch-independent MLP design.

Unlike the optical-flow and spatio-temporal modeling approaches discussed above, which incur substantial computational overhead from auxiliary modules,  this work introduces a lightweight solution that is inherently integrated into the model architecture.
We define an \emph{event} by means of a computationally trivial inter-frame pixel difference, and our update mechanism is seamlessly embedded within the MLP framework, operating without any external auxiliary models.

\section{Method}
\label{sec::method}

Our event-driven local update mechanism computes only the patches where events occur, which requires each patch to be processed independently.
As motivated in Section~\ref{sec::intro}, this patch independence is hard to obtain from the global receptive fields of CNNs or the all-to-all coupling of ViT attention, so we build evMLP entirely from MLPs.
This section presents the two components of our approach: the evMLP network architecture (Section~\ref{sec::method::net-arch}), which processes patches independently within each stage, and the event-driven local update mechanism (Section~\ref{sec::method::evupdate}), which exploits this property to skip unchanged patches during video processing.

\subsection{Network Architecture}
\label{sec::method::net-arch}

\begin{figure*}[ht]
    \centering
    \includegraphics[width=0.9\textwidth]{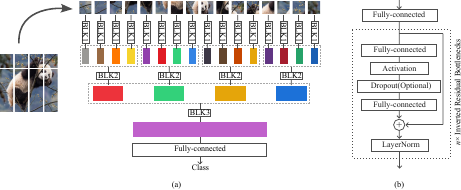}
    \caption{
        (a) shows the overview of the evMLP's architecture.
        The image or feature map is divided into fixed-size patches, and then each patch is processed independently using the proposed MLP-based building blocks.
        (b) shows the structure of the proposed MLP-based building block, which consists of a fully connected layer followed by $n$ Inverted Residual Bottlenecks.
        In each Inverted Residual Bottleneck, an activation function is applied after the first fully-connected layer, while the output of the second fully-connected layer undergoes a residual connection with the input data before being processed by a Layer Normalization operation for final output.
    }
    \label{fig::overview}
\end{figure*}

The overview of the proposed evMLP is illustrated in Fig.~\ref{fig::overview}(a). Given an image or feature map $\mathbf{x} \in \mathbb{R}^{H \times W \times C}$, we divide it into $N \times N$ patches of equal size.
Each patch is then flattened into a vector $\mathbf{x}_p \in \mathbb{R}^{C_{\text{in}}}$, where $p = 1, 2, \ldots, N^2$ and $C_{\text{in}} = (H/N) \times (W/N) \times C$.
These flattened patches are processed sequentially through the proposed MLP-based building block $\psi(\cdot)$, yielding the transformed outputs $\psi(\mathbf{x}_p) \in \mathbb{R}^{C_{\text{out}}}$.
By aggregating all processed patches, we obtain a new feature map $\mathbf{x}_{\{\psi(\mathbf{x}_p)\}} \in \mathbb{R}^{N \times N \times C_{\text{out}}}$.
In this way, each stage of the network independently processes every patch on the feature map, enabling selective computation only for patches where events occur.
While patches are processed independently within each stage, cross-patch communication is achieved across stages via the rearrange operation: the output patches of stage $l$ are reassembled into a new feature map, which is then re-divided into coarser patches for stage $l+1$. 
Over the six stages, the spatial resolution progresses from $224\!\times\!224$ through $32\!\times\!32$, $16\!\times\!16$, $8\!\times\!8$, $4\!\times\!4$, and $2\!\times\!2$ to a final $1\!\times\!1$ feature map, effectively growing the receptive field from local patches to the full image.

Fig.~\ref{fig::overview}(b) shows the structure of the proposed building block.
The process begins with a fully connected layer that mixes all input information and projects it to the specified dimension $C_{out}$, followed by $n$ Inverted Residual Bottlenecks.
Inspired by \cite{MobileNetV2, MobileNetV3}, we employ the inverted residual and bottleneck structure with two fully-connected layers in each Inverted Residual Bottleneck to reduce network parameters: the first layer expands the input $d_{in}$-dimensional features to $d_{out} = d_{in} \times \alpha$ dimensions through an expansion factor $\alpha$ followed by nonlinear activation, while the second layer projects the features back to $d_{in}$ dimensions and combines them with the input via residual connection before Layer Normalization.
In addition, an optional dropout can be added after each nonlinear activation function to prevent the network from overfitting.

\subsection{Event-Driven Local Update}
\label{sec::method::evupdate}

For a sequence of frames, the event-driven local update mechanism computes only the patches where events occur and reuses the previous frame's features for the remaining patches, thereby avoiding redundant computation.

Let $\mathbf{a}^l$ denote the current image/feature map to process, and $\mathbf{b}^l$ represent the image/feature map from the immediately preceding frame at stage $l$ (when $l$ is 1, $\mathbf{a}^1$ and $\mathbf{b}^1$ denote the raw image).
For an $n$-stage network, the image or feature map at the $l$-th stage is divided into ${N^l}^2$ patches of size $P^l \times P^l$ through a rearrange operation, resulting in $\{\mathbf{a}^l_p\}^{{N^l}^2}_{p=1}$ and $\{\mathbf{b}^l_p\}^{{N^l}^2}_{p=1}$.
The difference image $\mathbf{d}$ of the currently processed image $\mathbf{a}^1$ relative to the preceding image $\mathbf{b}^1$ is obtained via $\left|\mathbf{a}^1 - \mathbf{b}^1\right|$.
Considering the potential presence of noise from various sources in image sequences, we keep only the differences that reach the \emph{event threshold} $\tau$ and zero out the rest: $\mathbf{c}^0 = \mathbf{d} \odot \mathbf{1}_{\{\mathbf{d} \geq \tau\}}$, where $\mathbf{1}_{\{\cdot\}}$ is the element-wise indicator and $\odot$ the Hadamard product.
Subsequently, average pooling with a stride of $P^l$ is applied as $\mathbf{c}^l \leftarrow \text{AvgPool2D}(\mathbf{c}^{l-1}, P^l)$ to derive the event map $\mathbf{c}^l$.
Patches corresponding to non-zero pixels $\mathbf{c}^{l}_p$ in $\mathbf{c}^l$ indicate regions where events have occurred.
Therefore, the $l$-th stage $\psi^l(\cdot)$ of the network only needs to process the patches where events occur and update $\mathbf{a}^{l+1}_p $ with $ \psi^{l}(\mathbf{a}^l_p)$.
For patches where no event occurs, the corresponding feature $\mathbf{a}^{l+1}_p $ does not need to be calculated; instead, the result $\mathbf{b}^{l+1}_p $ from the previous frame can be directly reused.
By avoiding redundant calculations in this way, the computational efficiency can be improved.
The so proposed event-driven local update mechanism is summarized in Algorithm \ref{alg::evupdate}.

\begin{algorithm}
\caption{Event-driven local update mechanism}
\label{alg::evupdate}

\begin{algorithmic}[1]
\State $\mathbf{d} \leftarrow \left|\mathbf{a}^1 - \mathbf{b}^1 \right| $
\State $\mathbf{c}^0 = \mathbf{d} \odot \mathbf{1}_{\{\mathbf{d} \geq \tau\}}$
\For {$l = 1$ \textbf{to} $n$}
	\State $\{\mathbf{a}^l_p\}^{{N^l}^2}_{p=1} \leftarrow \mathbf{a}^l $,
	$\{\mathbf{b}^l_p\}^{{N^l}^2}_{p=1} \leftarrow \mathbf{b}^l $
	\State $\mathbf{c}^{l} \leftarrow AvgPool2D(\mathbf{c}^{l-1}, P^l) $	
	\For {$p=1$ \textbf{to} ${N^{l}}^2$ }
		\If {$\mathbf{c}^{l}_p \neq \mathbf{0} $}
			\State $\mathbf{a}^{l+1}_p \leftarrow \psi^{l}(\mathbf{a}^l_p)$
		\Else
			\State $\mathbf{a}^{l+1}_p \leftarrow \mathbf{b}^{l+1}_p$
		\EndIf
	\EndFor
	\State $\mathbf{a}^{l+1} \leftarrow \{\mathbf{a}^{l+1}_p\}^{{N^l}^2}_{p=1}$
\EndFor

\end{algorithmic}

\end{algorithm}

A subtlety of the frame-to-frame comparison in Algorithm~\ref{alg::evupdate} is that a patch whose per-frame change consistently falls below $\tau$ is never updated, allowing small errors to accumulate gradually over long sequences. This cumulative drift is a known limitation of per-frame thresholding and is discussed further in Section~\ref{sec::limitations}.

\section{Experiments}
\label{sec::experiments}

We evaluate the performance of the proposed evMLP on image classification tasks, and demonstrate the efficacy of our event-driven local update mechanism through computational cost analysis in video processing.
Controlled experiments further examine how event threshold settings trade off output consistency against computational efficiency.

\subsection{Implementation Details}
\label{sec::exp::impl}

Following the architecture in Section \ref{sec::method::net-arch}, we implement evMLP for image classification.
The network takes $224\!\times\!224\!\times\!3$ inputs, processes them through 6 stages of our evMLP's building block to generate $1\!\times\!1\!\times\!512$ feature maps, and produces $k$-dimensional outputs via a fully-connected layer.
We employ GELU\cite{GELU} activation functions, with dropout applied to randomly zero features at tuned probabilities for mitigating overfitting.
Configurations are detailed in Table \ref{table::net_cfg}.
These per-stage dimensions and depths are a deliberately simple instantiation rather than the outcome of architecture search; because the event-driven mechanism (Section~\ref{sec::method::evupdate}) relies only on patch independence, it applies unchanged to any such configuration.

\begin{table}[ht]

\caption{
    Specifications for evMLP.
    BLK denotes our evMLP's building block described in Section \ref{sec::method::net-arch}.
    FC denotes fully-connected.
    $P$ denotes the patch size, $\alpha$ denotes the expansion factor, $d_{out}$ denotes the output dimension, and $n$ denotes the number of residual blocks in the BLK.
}
\label{table::net_cfg}
\center
\small
\begin{tabular}{l c c c c c c}
  \toprule
  Input               & Block & $P$ & $\alpha$ & $C_{out}$ & $n$ & Dropout \\
  \midrule
  $224\!\times\!224\!\times\!3$  & BLK &  7  &  4  &    64     & 5   & 0   \\
  $ 32\!\times\!32\!\times\!64$  & BLK &  2  &  4  &   128     & 5   & 0   \\
  $ 16\!\times\!16\!\times\!128$ & BLK &  2  &  2  &   512     & 7   & 0   \\
  $ 8\!\times\!8\!\times\!512$   & BLK &  2  &  2  &   512     & 7   & 0   \\
  $ 4\!\times\!4\!\times\!512$   & BLK &  2  &  2  &   512     & 9   & 0   \\
  $ 2\!\times\!2\!\times\!512$   & BLK &  2  &  2  &   512     & 9   & 0.2 \\
  $ 1\!\times\!1\!\times\!512$   & FC  &  -  &  -  &   $k$     & -   & -   \\

  \bottomrule
\end{tabular}
\end{table}

\subsection{ImageNet Classification}
\label{sec::exp::imgnet}

We evaluated our evMLP on the ImageNet-1K\cite{ImageNet1K} dataset, which contains a training set of 1.28 million images across 1000 categories and a validation set of 50,000 images.
We train our model from scratch using PyTorch\cite{PyTorch} with the SGD optimizer with momentum.
We use a momentum of 0.9, weight decay of $1e^{-5}$, batch size of 1024, and initial learning rate of 0.1.
We employed a cosine scheduler to decay the learning rate over 300 epochs, incorporating a linear warm-up for the first 5 epochs.
Augmentation and regularization strategies similar to those in \cite{ViT,DeiT,MLPMixer} were adopted, including auto-augment, label smoothing, mixup, cutmix, and random erasing.

\begin{table}[ht]

\caption{
    Performance comparisons of various models on the ImageNet-1K validation set, all models are trained from scratch without extra data.
    Models above the middle rule require lower computational cost, while those below incur higher computational cost.
    Throughput is measured on an NVIDIA RTX 4090 GPU, using the GitHub repository provided in \cite{PyTorchImageModels}.
}
\label{table::result_imgnet}
\center
\small
\setlength{\tabcolsep}{4pt}
\begin{tabular}{l | c c c c }
  \toprule
  Model                  &\makecell[c]{MACs\\(G)} &  \makecell[c]{Params\\(M)} &  \makecell[c]{Throughput\\(imgs/sec)}& \makecell[c]{Top-1\\Acc}\\
  \midrule
  ResNet18\cite{ResNet}     & 1.82  & 11.69 & 8293           &  69.7\%           \\
  T2T-ViT-7\cite{T2T}       & 1.16  & 4.31 & 4657           &  71.7\%           \\
  DeiT-Ti\cite{DeiT}        & 1.26  &  5.71  & 9962           &  72.2\%           \\
  gMLP-Ti\cite{gMLP}        & 1.35  &  5.88  & 4213           &  72.3\%           \\
  \textbf{evMLP}            & 1.03  &  38.42 & \textbf{11491} &  \textbf{73.5\% } \\
  \midrule
  ViT-B/16\cite{ViT}        & 17.58 &  86.57 & 1493           &  79.6\%           \\
  Mixer-B/16\cite{MLPMixer} & 12.62 &  59.88 & 1674           &  76.4\%           \\
  ResMLP-B24\cite{ResMLP}   & 23.02 &  115.74 & 864            &  81.0\%           \\
  \bottomrule
\end{tabular}
\end{table}

We compare the classification performance of our evMLP with several mainstream state-of-the-art architectures on the ImageNet-1K validation set.
Given the relatively low computational cost of our evMLP model, we compare it with models including ResNet-18\cite{ResNet}, T2T-ViT-7\cite{T2T}, DeiT-Ti\cite{DeiT} and gMLP-Ti\cite{gMLP}, which have similar complexity (measured by MACs).
For reference, we also include higher-computational-cost models ViT-B/16\cite{ViT} and Mixer-B/16\cite{MLPMixer}.
All models are trained from scratch without extra data with $224\!\times\!224$ input resolution.
The comparison results are shown in Table \ref{table::result_imgnet}.
Experimental results demonstrate that our evMLP model achieves the highest top-1 accuracy of 73.5\% at a computational cost of 1.03 GMACs, under comparable levels of computational cost.
Additionally, it remains competitive with models exceeding 10 GMACs in computational cost.

\noindent
\textbf{Model throughput} We also evaluated the throughput of all models on a single NVIDIA RTX 4090 GPU using the code provided in \cite{PyTorchImageModels}.
Due to the patch-wise independent computation design of our evMLP, the parallelism of its inference is greatly enhanced.
As shown in Table \ref{table::result_imgnet}, our model consequently achieves the highest throughput of 11,491 images/sec, significantly surpassing other models of comparable computational cost.
Beyond parallelism, every stage reduces to a regular, dense matrix multiplication, avoiding the irregular memory access of convolutions (im2col) and the quadratic token mixing of attention; this makes evMLP well matched to the BLAS-optimized datapaths and cache hierarchies of modern hardware, and is a second reason for its throughput advantage.

\begin{figure}[ht]
    \centering
    \includegraphics[width=0.46\textwidth]{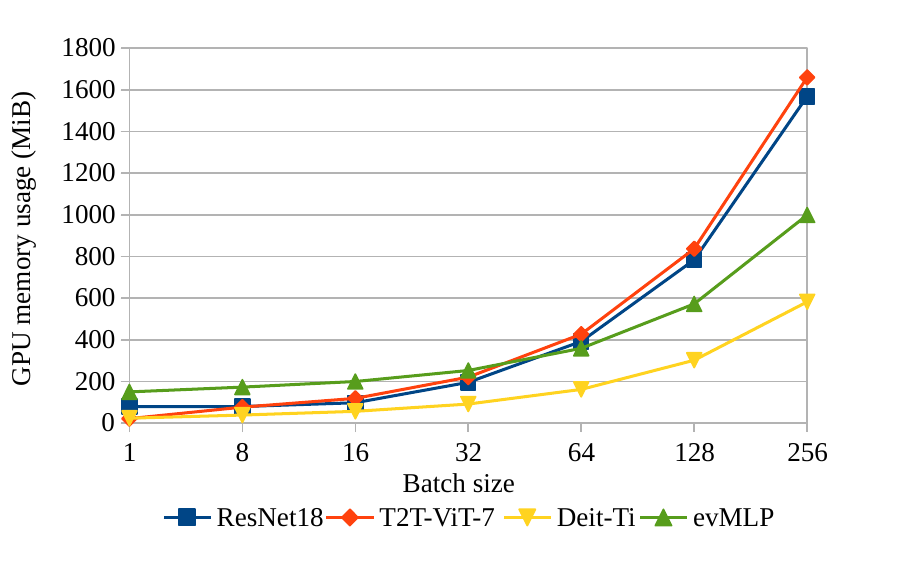}
    \caption{
        Comparison of GPU memory consumption across different models under varying batch sizes.
        Inference was performed using NVIDIA TensorRT\cite{trt}.
    }
    \label{fig::gpu-mem}
\end{figure}

\noindent
\textbf{Memory usage} As shown in Table \ref{table::result_imgnet}, our evMLP has a computational cost of only 1.03 GMACs, yet its parameter count reaches 38.42M, substantially higher than models of comparable MACs. This is a consequence of the hierarchical patch-wise design: the high-dimensional stages ($C_{out}=512$) with deep inverted residual blocks ($n=7$--$9$ per stage) accumulate parameters, while the computational cost per patch remains low due to the small patch size ($P=2$) in these stages.
However, a larger number of model parameters does not necessarily imply a higher memory overhead during inference, as the inference memory footprint includes both model parameters and intermediate activation caches (e.g., feature maps).
Consequently, as the batch size fed into the model increases, the reuse efficiency of model parameters also improves.
We conducted inference using NVIDIA TensorRT\cite{trt} to measure GPU memory usage across different batch sizes.
The experimental results are illustrated in Fig.~\ref{fig::gpu-mem}.
When the batch size is less than 32, our evMLP consistently exhibits the highest memory usage.
However, when the batch size reaches 64 or larger, our evMLP begins to demonstrate an advantage in memory consumption, being significantly lower than ResNet18 and T2T-ViT-7, and only slightly higher than DeiT-Ti.
In other words, evMLP trades parameter count for activation efficiency: the parameters are a fixed, read-only cost, whereas the per-input activation footprint, which dominates at practical batch sizes, stays small, which is why its memory usage falls below comparable models once the batch is large enough.

\noindent
\textbf{Knowledge distillation} Following the hard-label distillation approach in DeiT\cite{DeiT}, we employ an EfficientNetV2-S\cite{EfficientNetV2} model as the teacher model to distill our evMLP student model on the ImageNet-1K training set, which achieves 81.3\% top-1 accuracy on ImageNet-1K at $224\!\times\!224$ resolution.
We compare the performance of models with similar computational costs (in MACs) after knowledge distillation, with the results presented in Table \ref{table::result_imgnet_distillation}.
Experimental results show that our evMLP benefits from knowledge distillation with a CNN teacher, reaching 75.4\% top-1 accuracy.
Although we intentionally selected a relatively compact teacher model slightly less powerful than the RegNetY-16GF (82.9\% top-1) used in \cite{T2T,DeiT}, the distilled version of our model still outperforms T2T-ViT-7 (73.1\%) and DeiT-Ti (74.5\%) in top-1 accuracy.
The upper block of Table~\ref{table::result_imgnet_distillation} follows the same 300-epoch schedule as the baselines, ensuring a controlled comparison at a comparable training budget. The last row reports a single extended run: the distilled evMLP trained for 800 epochs (denoted evMLP$_{800\mathrm{ep}}$), which raises its top-1 accuracy to 77.0\%. We emphasize that this entry is deliberately \emph{not} directly comparable to the others, which use a shorter schedule, and we do not use it to claim any advantage over prior work; it is included only as an indication of the architecture's headroom. The gain is consistent with the well-documented tendency of all-MLP architectures to converge slowly and to keep improving under longer training and stronger augmentation~\cite{MLPMixer,ResMLP}, suggesting that evMLP's accuracy at 300 epochs is not saturated and that the architecture has meaningful untapped potential.

\begin{table}[ht]

\caption{
    Performance comparison of different models with distillation.
    An alembic symbol (\includegraphics[width=0.6em]{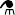}) indicates models trained with distillation.
    We use an EfficientNetV2-S\cite{EfficientNetV2} model with a top-1 accuracy of 81.3\% on ImageNet-1K at $224\!\times\!224$ resolution as the teacher.
    The upper block uses a controlled 300-epoch schedule for a fair comparison with the baselines. The last row (evMLP trained for 800 epochs) is reported only as an indication of the architecture's headroom and is \emph{not} directly comparable to the other entries.
}

\label{table::result_imgnet_distillation}
\center
\small
\setlength{\tabcolsep}{4pt}
\begin{tabular}{l | c c c c }
  \toprule
  Model                  &\makecell[c]{MACs\\(G)} &  \makecell[c]{Params\\(M)} & \makecell[c]{Epochs} & \makecell[c]{Top-1\\Acc}\\
  \midrule
  T2T-ViT-7 \includegraphics[width=0.6em]{pics/sym-distillation.pdf}      & 1.16  &  4.31  & 310 &  73.1\%           \\
  DeiT-Ti \includegraphics[width=0.6em]{pics/sym-distillation.pdf}        & 1.26  &  5.71  & 300 &  74.5\%           \\
  \textbf{evMLP \includegraphics[width=0.6em]{pics/sym-distillation.pdf}} & 1.03  &  38.42 & 300 &  \textbf{75.4\% } \\
  \midrule
  \textbf{evMLP$_{800\mathrm{ep}}$ \includegraphics[width=0.6em]{pics/sym-distillation.pdf}} & 1.03  &  38.42 & 800 &  \textbf{77.0\% } \\
  \bottomrule
\end{tabular}
\end{table}

\begin{table}[ht]
\caption{evMLP vs.\ the parameter-efficient variant evMLP-C on ImageNet-1K (trained from scratch, 300 epochs, identical recipe). Throughput is measured on an RTX 4090. The last column reports the $\tau=0$ event-driven MACs saving on the UCF-Crime timing subset (same protocol as Section~\ref{sec::exp::wallclock}).}
\label{table::nconv}
\center
\small
\setlength{\tabcolsep}{4pt}
\begin{tabular}{l c c c c c}
  \toprule
  Model & \makecell[c]{MACs\\(G)} & \makecell[c]{Params\\(M)} & \makecell[c]{Throughput\\(imgs/sec)} & \makecell[c]{Top-1\\Acc} & \makecell[c]{$\tau{=}0$ MACs\\saving} \\
  \midrule
  evMLP   & 1.03 & 38.42 & 11491 & 73.5\% & 27.0\% \\
  evMLP-C & 1.03 & 12.40 &  9140 & 73.8\% & 26.2\% \\
  \bottomrule
\end{tabular}
\end{table}

\noindent
\textbf{Parameter efficiency} The 38.42M parameter count is a property of this particular all-MLP instantiation, not of the event-driven mechanism. As an existence proof we evaluate evMLP-C, which retains the first three of evMLP's six MLP stages (reducing the input to an $8\times8\times512$ feature map) and replaces the remaining three with a lightweight convolutional head: a $1\times1$ channel projection, a stack of inverted residual blocks~\cite{MobileNetV2} (with GELU activations rather than the original ReLU6), a $1\times1$ channel expansion, global average pooling, and the linear classifier. The split is deliberate. The retained MLP stages preserve patch independence and therefore carry the event-driven skipping, while the convolutional tail raises parameter efficiency on the coarse stages where skipping contributes least. These are standard components; we do not tune the head, and its exact configuration is in the released code. At the same 1.03 GMACs, evMLP-C reduces the parameter count by $3.1\times$ (38.42M to 12.40M) and, trained identically from scratch, slightly improves top-1 accuracy (Table~\ref{table::nconv}), at a moderate cost in inference throughput. We analyse this trade-off in Section~\ref{sec::limitations}. We report evMLP-C only to show that the mechanism transfers to parameter-efficient designs; the backbone itself is not a contribution of this work.

\subsection{Computational Cost Analysis for Video Processing}
\label{sec::exp::cost}

Video processing provides a natural setting to validate the event-driven local update mechanism, as consecutive frames let it skip patches without events.
Note that in this paper evMLP is evaluated as a frame-level classifier; the video datasets here serve solely as sources of natural frame sequences for validating the event-driven mechanism's computational savings, not for evaluating action recognition or video understanding tasks.
We process videos in the video datasets HMDB51\cite{HMDB51}, MSVD\cite{MSVD}, UCF101\cite{UCF101}, and UCF-Crime\cite{UCFCRIME} using our evMLP (for UCF-Crime, only the testing set is utilized), comparing the computational performance of the model with and without the event-driven local update mechanism.
We set the event threshold $\tau$ (refer to Section \ref{sec::method::evupdate}) to 0, meaning any pixel change is considered an event occurrence.
This ensures that using the event-driven local update mechanism does not affect the processing results, allowing us to focus solely on changes in computational cost.
Frames in the videos are resized to $224\!\times\!224$ and processed frame-by-frame.
We report and compare the average MACs per frame processed by our evMLP model with and without the event-driven local update mechanism across multiple datasets, as shown in Table \ref{table::result_vid_proc}.

\begin{table}[ht]

\caption{
    Comparison of computational cost (MACs(G)/frame) with/without the event-driven local update mechanism across video datasets.
    \texttimes \ denotes the dense baseline (mechanism disabled), while \checkmark \ represents the case with the mechanism enabled, where the event threshold $\tau = 0$.
}
\label{table::result_vid_proc}
\center
\small
\begin{tabular}{l l l}
  \toprule
       & evMLP$^{\times}$ & evMLP$^{\checkmark}$ \\
  \midrule
  HMDB51\cite{HMDB51}      & 1.030 &  0.882 ${(\downarrow 14.4\%)}$ \\
  MSVD\cite{MSVD}          & 1.030 &  0.943 ${(\downarrow 8.4\%)}$  \\
  UCF101\cite{UCF101}      & 1.030 &  0.881 ${(\downarrow 14.5\%)}$ \\
  UCF-Crime\cite{UCFCRIME} & 1.030 &  0.754 ${(\downarrow 26.8\%)}$ \\
  \bottomrule
\end{tabular}
\end{table}

Experimental results show that our proposed event-driven local update mechanism improves the computational performance of our evMLP across diverse video datasets.
For Action Recognition datasets HMDB51 and UCF101 (relatively simple scenes), computational cost decreased by 14.4\% and 14.5\%, reducing the average per-frame cost from 1.030 GMACs to 0.882 GMACs and 0.881 GMACs.
For Video Description dataset MSVD (richer scenes), the improvement was more modest with a 8.4\% computational cost reduction, yielding an average of 0.943 GMACs per frame.
On surveillance dataset UCF-Crime (stationary cameras, reduced camera-motion \emph{events}), computational performance significantly improves, with a 26.8\% reduction in average per-frame computational cost, achieving an average of 0.754 GMACs per frame.

\subsection{Study of the Event Threshold}
\label{sec::exp::evTh}

Consecutive video frames often exhibit noise that is semantically irrelevant to image content.
Under stringent event thresholds, however, corresponding patches may be erroneously flagged as containing events and consequently updated.
Moreover, by virtue of deep neural networks' robust generalization capabilities, semantically insignificant variations may exert negligible influence on the classification outputs.
While experiments in Section \ref{sec::exp::cost} validate the efficacy of the event-driven local update mechanism, conservatively configuring the event threshold at $\tau=0$, albeit ensuring output consistency with the dense baseline, fails to yield substantial computational efficiency improvements.
In this section we study the impact of event threshold configurations on evMLP's computational efficiency and output consistency.

\begin{figure}[ht]
    \centering
    \begin{minipage}[t]{0.19\linewidth}
    \includegraphics[width=1.0\textwidth]{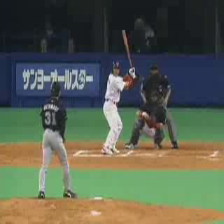}
    \includegraphics[width=1.0\textwidth]{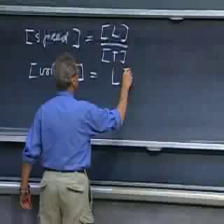}
    \centerline{\footnotesize (a)}
    \end{minipage}
    \begin{minipage}[t]{0.19\linewidth}
    \includegraphics[width=1.0\textwidth]{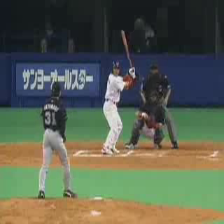}
    \includegraphics[width=1.0\textwidth]{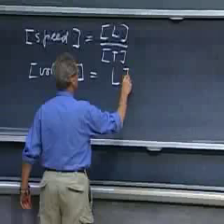}
    \centerline{\footnotesize (b)}
    \end{minipage}
    \begin{minipage}[t]{0.19\linewidth}
    \includegraphics[width=1.0\textwidth]{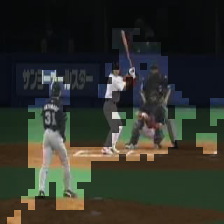}
    \includegraphics[width=1.0\textwidth]{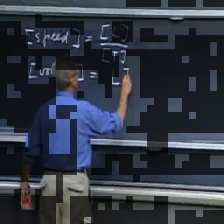}
    \centerline{\footnotesize (c) $\tau=0.05$}
    \end{minipage}
    \begin{minipage}[t]{0.19\linewidth}
    \includegraphics[width=1.0\textwidth]{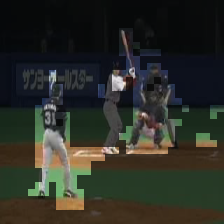}
    \includegraphics[width=1.0\textwidth]{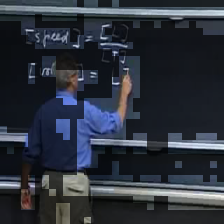}
    \centerline{\footnotesize (d) $\tau=0.10$}
    \end{minipage}
    \begin{minipage}[t]{0.19\linewidth}
    \includegraphics[width=1.0\textwidth]{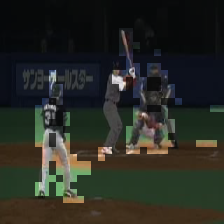}
    \includegraphics[width=1.0\textwidth]{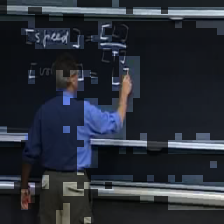}
    \centerline{\footnotesize (e) $\tau=0.15$}
    \end{minipage}
    \caption{
        Comparison of event maps generated with different event thresholds.
        (a), (b) are consecutive video frames.
        (c)-(e) are results of event maps generated using different event thresholds $\tau$ and superimposed on frame (b), brighter areas indicate patches where events occur.
    }
    \label{fig::evThDemo}
\end{figure}

In the experiments, all input images were preprocessed using the standard normalization procedure commonly adopted in the literature.
Specifically, pixel values were first scaled to the range [0, 1] and then normalized channel-wise using the mean values [0.485, 0.456, 0.406] and standard deviations [0.229, 0.224, 0.225] derived from the ImageNet-1K\cite{ImageNet1K} dataset.
Fig.~\ref{fig::evThDemo} visualizes event maps generated under different event thresholds, demonstrating that aggressive threshold settings significantly reduce event regions, thereby improving computational performance, whereas excessively large thresholds may neglect authentic events, potentially compromising processing outcomes.
When the event threshold $\tau$ is set to 0, evMLP yields results identical to the dense baseline.
Therefore, we establish the frame-by-frame processing results at $\tau=0$ as the reference baseline.
This enables quantitative evaluation of event threshold impact by comparing consistency rates of different thresholds against the $\tau=0$ baseline.
Here, consistency denotes the fraction of processed frames whose top-1 prediction matches that of the $\tau=0$ baseline.

\begin{figure}[ht]
	\centering
    \includegraphics[width=0.46\textwidth]{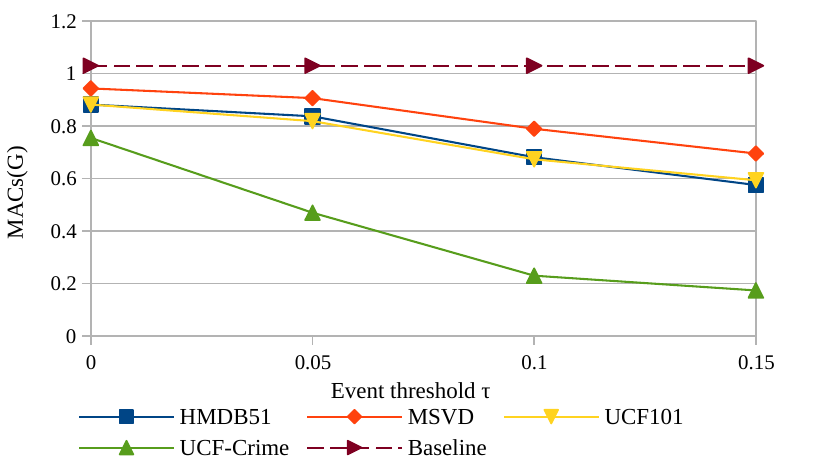}\\
    {\footnotesize (a) Impact of event thresholds on computational cost across datasets. The dense baseline is plotted as a dashed line.}
    \includegraphics[width=0.46\textwidth]{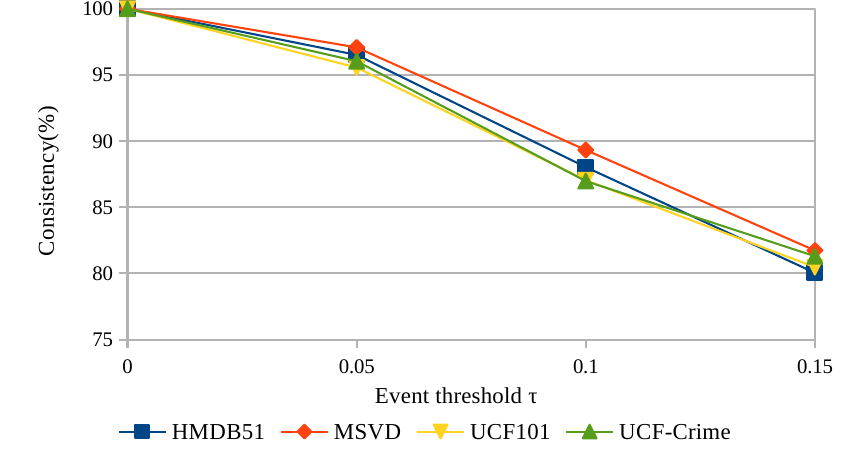}\\
	{\footnotesize (b) Impact of event thresholds on model consistency with the $\tau=0$ baseline across datasets.}
    \caption{Impact of event thresholds on output consistency and computational cost across datasets,  $\tau \in \{0, 0.05, 0.1, 0.15\}$.}
    \label{fig::evThVs}
\end{figure}

We apply the non-distilled version of the evMLP model, pre-trained on ImageNet-1K as described in Section \ref{sec::exp::imgnet} to the same video datasets as in Section~\ref{sec::exp::cost} with varying event thresholds.
During experimentation, video frames are resized to $224\!\times\!224$ and fed into the model to produce classification outputs.
As shown in Fig.~\ref{fig::evThVs} and Table~\ref{table::consistency}, varying event thresholds ($\tau \in \{0, 0.05, 0.1, 0.15\}$) significantly impact both per-frame computational cost (GMACs/frame) and output consistency across all evaluated datasets.

\begin{table}[ht]
\caption{Top-1 consistency (\%) with the $\tau=0$ baseline across datasets under varying event thresholds. These values accompany Fig.~\ref{fig::evThVs}(b); the corresponding per-frame computational cost is shown in Fig.~\ref{fig::evThVs}(a).}
\label{table::consistency}
\center
\small
\begin{tabular}{c c c c c}
  \toprule
  $\tau$ & HMDB51 & MSVD & UCF101 & UCF-Crime \\
  \midrule
  0.00 & 100.0 & 100.0 & 100.0 & 100.0 \\
  0.05 &  96.5 &  97.1 &  95.6 & 96.0 \\
  0.10 &  88.0 &  89.3 &  87.0 & 87.0 \\
  0.15 &  80.0 &  81.7 &  80.5 & 81.3 \\
  \bottomrule
\end{tabular}
\end{table}

Experimental results demonstrate that increasing the event threshold classifies more patches as event-inactive.
Under our event-driven local update mechanism, computation for these patches is skipped, thus reducing computational cost while correspondingly decreasing consistency with the $\tau=0$ baseline.
At $\tau = 0.05$, computational cost reductions exceeding 10\% are achieved across all datasets while maintaining consistency with the $\tau=0$ baseline above 90\%.
When $\tau$ increases to 0.15, aggressive event thresholds yield significant computational savings surpassing 30\% on all datasets, yet reduce consistency to around 80\%.
Additionally, as plotted, our evMLP featuring the event-driven local update mechanism shows significant performance differences across datasets.

The results  exhibit substantial performance variations across datasets.
As HMDB51\cite{HMDB51} and UCF101\cite{UCF101} are action recognition datasets, results on these datasets show similar computational performance and consistency trends under varying event thresholds.
The results on UCF-Crime~\cite{UCFCRIME} show a marked improvement in computational performance, whereas the improvement observed on MSVD~\cite{MSVD} is more conservative.
This prompted a focused analysis of the scenarios in UCF-Crime and MSVD to investigate this discrepancy.

The stationary camera recordings avoid camera-motion artifacts, enabling our evMLP to achieve significantly superior computational performance on UCF-Crime at identical event thresholds compared to other datasets.
At $\tau = 0.05$, the average per-frame computational cost reduces to 0.470 GMACs/frame, less than half the baseline (1.030 GMACs/frame) while maintaining a consistency with the $\tau=0$ baseline of 96.0\%.
With increasing $\tau$, the improvement in computational performance on dataset UCF-Crime remains significantly ahead of that on other datasets.
However, excessive update skipping substantially compromises output consistency despite the improved computational efficiency.
At $\tau=0.15$, computational cost on UCF-Crime reduces by over 80\% to an average of 0.173 GMACs/frame, whereas consistency drops to 81.3\%.
Results on dataset MSVD exhibit more conservative performance gains.
As a video description dataset featuring diverse scenes and moving cameras, MSVD generates frequent inter-frame events.
At $\tau = 0.05$, our evMLP achieves merely 12\% computational improvement on it.
Although MSVD achieves lower computational savings than other datasets, it maintains higher consistency at identical thresholds, as its frequent inter-frame events cause fewer patches to be skipped. At $\tau = 0.15$ it reduces computational cost to 0.695 GMACs/frame while diminishing consistency to 81.7\%.

In summary, the efficiency of our event-driven mechanism is highly context-dependent: it delivers the most pronounced gains on surveillance-type datasets with stationary cameras (UCF-Crime) by leveraging background stability, while the benefits are more measured on datasets characterized by moving cameras and dynamic scenes (MSVD).
This clearly defines the optimal application domain for our approach.

\subsection{Wall-Clock Latency Analysis}
\label{sec::exp::wallclock}

The preceding analysis quantifies efficiency in MACs, which reflect arithmetic and energy cost.
Here we examine whether these savings translate into wall-clock latency reduction on general-purpose hardware, using the surveillance dataset UCF-Crime where the mechanism is most effective.
This concerns the per-frame latency of the event-driven \emph{mechanism}, a separate question from the dense architecture's large-batch throughput (Table~\ref{table::result_imgnet}): strong throughput there does not by itself guarantee that the sparsity savings become wall-clock gains here.
Since wall-clock speedup is approximately proportional to MACs reduction in the compute-bound (single-core) regime, the results on UCF-Crime represent an upper bound on the speedup achievable on the other datasets, whose MACs savings are lower (Table~\ref{table::result_vid_proc}).
For timing we cap each video at its first 500 frames (73{,}016 frames in total), as the testing videos vary enormously in length (from a few hundred to over 100{,}000 frames) and the longest would otherwise dominate the wall-clock time and unbalance multi-stream processing.
This subset preserves the property that governs timing: its MACs reduction (27.0\% at $\tau=0$) closely matches the full dataset (26.8\%).

\noindent\textbf{Single-core performance.}
On a single CPU thread the execution is compute-bound, and the event-driven mechanism delivers speedups that closely track its MACs savings.
\setlength{\tabcolsep}{4pt}
\begin{table}[ht]
\caption{Wall-clock speedup under varying event thresholds on a single CPU core (Xeon E5-2699v4, UCF-Crime timing subset of 73{,}016 frames). MACs saved is measured on the same subset and closely tracks the full-dataset values; the corresponding output consistency is a hardware-independent quantity reported on the full dataset in Section~\ref{sec::exp::evTh}.}
\label{table::wallclock_single}
\center
\small
\begin{tabular}{c c c}
  \toprule
  $\tau$ & MACs saved & Speedup \\
  \midrule
  0.00 & 27.0\% & 1.26$\times$ \\
  0.05 & 54.6\% & 1.68$\times$ \\
  0.10 & 77.9\% & 2.62$\times$ \\
  0.15 & 82.7\% & 3.00$\times$ \\
  \bottomrule
\end{tabular}
\end{table}
As shown in Table~\ref{table::wallclock_single}, at $\tau=0$ the per-frame speedup reaches 1.26$\times$, closely approaching the theoretical upper bound implied by the 27.0\% MAC reduction.
Increasing $\tau$ trades output consistency for speed: at $\tau=0.15$ the speedup rises to 3.00$\times$, with the corresponding consistency on the full UCF-Crime testing set (81.3\%) reported in Section~\ref{sec::exp::evTh}.

\noindent\textbf{Multi-thread scaling.}
However, this correspondence breaks down when parallelism increases.
Because the event-driven mechanism issues the same number of operators as a dense pass, only with smaller operands, and adds gather/scatter for active patches, the fixed per-operator overhead (dispatch, allocation, and thread synchronization) does not shrink with the arithmetic.
Naively increasing intra-op thread count therefore erodes the benefit, as the parallelizable compute is divided across threads while the fixed overhead is not.
This motivates the stream-level deployment strategy described below.

\noindent\textbf{Stream-level parallelism.}
The streaming, single-frame setting that motivates the event-driven mechanism is common in practice, where multiple independent video sources are processed concurrently (e.g., multi-camera surveillance).
Although each stream processes one frame at a time, the model weights are read-only and shared across all streams, so the memory footprint does not grow linearly with the number of streams, akin to increasing the batch size in a single-stream setting.
The correct deployment strategy for such multi-core systems is stream-level rather than operator-level parallelism.
We assign one single-threaded inference worker to each video stream and run as many workers as there are cores, so each stream retains its near-ideal per-frame speedup.
On a 22-core CPU the system throughput scales to 343.8 FPS at 22 streams, a 16.8\% improvement over the dense baseline at the same stream count (294.4 FPS).
As shown in Table~\ref{table::multistream}, the per-stream speedup stays in the range of 1.17$\times$ to 1.26$\times$ across all stream counts, confirming that stream-level parallelism preserves the per-frame benefit that operator-level parallelism erodes.
\setlength{\tabcolsep}{4pt}
\begin{table}[ht]
\caption{Multi-stream scaling on a 22-core CPU (Xeon E5-2699v4, DDR4-2400, $\tau=0$). Each stream uses one thread, and the model is stored once and shared read-only across all streams.}
\label{table::multistream}
\center
\small
\begin{tabular}{c c c c}
  \toprule
  Streams & Dense (FPS) & Event (FPS) & Per-stream speedup \\
  \midrule
   1 &  18.1 &  22.8 & 1.26$\times$ \\
   2 &  36.2 &  45.7 & 1.26$\times$ \\
   4 &  73.2 &  91.3 & 1.25$\times$ \\
   8 & 142.4 & 175.2 & 1.23$\times$ \\
  16 & 257.2 & 314.2 & 1.22$\times$ \\
  22 & \textbf{294.4} & \textbf{343.8} & 1.17$\times$ \\
  \bottomrule
\end{tabular}
\end{table}
Scaling is near-linear through the first 8 cores and keeps rising across all 22 physical cores, peaking at 22 streams (event 343.8 FPS, dense 294.4 FPS) with no regression. Storing the model once and sharing its read-only weights across streams avoids the per-stream weight replication that would otherwise dominate memory traffic, pushing the memory-bandwidth wall beyond this platform's 22-core budget. The per-stream speedup eases from 1.26$\times$ to 1.17$\times$ as streams are added, indicating the workload approaches this bandwidth wall without reaching it on this platform. This decline reflects bandwidth contention rather than reduced turbo frequency: at a given stream count dense and event activate the same number of cores and run at the same clock, so frequency cancels in their ratio, whereas event, which computes fewer MACs yet reads the same weights (a lower arithmetic intensity), feels the growing bandwidth pressure first. This sensitivity is intrinsic to evMLP, whose 1.03 GMACs are spread over 38.42M weights: at large batch sizes the weights amortize across the batch, so the pressure is hidden and the regular dense GEMMs deliver the high throughput of Table~\ref{table::result_imgnet}, whereas in single-frame streaming the weights are re-read every frame, so aggregate weight bandwidth is what ultimately bounds concurrency on general-purpose processors, whose datapaths favor dense, regular computation over sparse, data-dependent workloads.

\noindent\textbf{Discussion.}
Our GPU evaluation covers the dense architecture (large-batch throughput in Table~\ref{table::result_imgnet}, memory usage in Fig.~\ref{fig::gpu-mem}); we do not report GPU wall-clock latency for the event-driven mechanism itself, and we clarify why.
Beyond the batch-size regime difference noted above, the mechanism transforms dense, regularly-shaped computation into data-dependent, irregular operations (variable-size matrix multiplications together with gather/scatter over the active patches).
On a massively parallel processor a lightweight model is already throughput-bound rather than compute-bound, so the fixed per-kernel launch overhead and irregular memory access prevent MACs reduction from translating into latency reduction.
Reporting a near-unity GPU speedup would conflate a mismatch between hardware and workload with the efficacy of the mechanism itself.

The CPU single-core results above confirm that the mechanism's arithmetic savings do translate into wall-clock improvement in a compute-bound regime; realizing its full potential calls for dataflow-oriented hardware, which we discuss in Section~\ref{sec::future}.

\subsection{Limitations}
\label{sec::limitations}

We make several limitations explicit. First, evMLP's parameter count (38.42M) is higher than models of comparable MACs, a consequence of the fully-connected layers within each patch lacking weight sharing. While this count is moderate by modern standards, and the inference memory footprint is dominated by intermediate features rather than parameters at practical batch sizes (Fig.~\ref{fig::gpu-mem}), it does contribute to the memory bandwidth pressure observed in multi-stream deployment (Section~\ref{sec::exp::wallclock}). This parameter count is not fixed by the mechanism and can be reduced in practice: the evMLP-C variant (Section~\ref{sec::exp::imgnet}) replaces the coarse final stages with a convolutional head and, as reported in Table~\ref{table::nconv}, uses $3.1\times$ fewer parameters (12.40M) at the same MACs and comparable accuracy. This comes with two costs we measure, both because the convolutional head is not patch-independent: it cannot be event-skipped and is always computed densely, so its $\tau=0$ MACs saving on the UCF-Crime subset is 26.2\% rather than evMLP's 27.0\% (Table~\ref{table::nconv}); and its convolutions are less suited to the dense matrix multiplications that make evMLP fast, lowering throughput to about 79\% of evMLP's. evMLP-C is thus a practical way to relieve the parameter cost when that is the binding constraint, whereas evMLP remains preferable when throughput is the priority. Second, the frame-to-frame comparison in Algorithm~\ref{alg::evupdate} means that a patch whose per-frame change consistently falls below $\tau$ is never updated, allowing small errors to accumulate gradually over long sequences. Third, the computational benefit of the event-driven mechanism is strongly scene-dependent: it is greatest for stationary-camera footage with stable backgrounds and considerably smaller for dynamic scenes with camera motion, as our cross-dataset results demonstrate. Fourth, the event-driven mechanism's wall-clock behavior on GPUs remains unverified; as discussed in Section~\ref{sec::exp::wallclock}, this reflects a mismatch between hardware and workload rather than a limitation of the mechanism itself. None of these limitations undermines the core contribution, and several of them directly motivate the dedicated hardware realization discussed below.

\subsection{Future Work}
\label{sec::future}

Several directions are worth exploring. The most natural next step is a dedicated hardware realization of evMLP on FPGA or ASIC, where change detection and sparse patch activation can be implemented natively in the datapath, allowing the arithmetic savings to translate directly into latency and energy reduction. On the algorithmic side, the slow drift in frame-to-frame thresholding could be mitigated by periodic full-frame refreshes or adaptive threshold schedules.
Having shown with evMLP-C that a convolutional head recovers most of the efficiency at a fraction of the parameters, a systematic study of such hybrid MLP-convolution designs is a natural next step, including where to place the split and how it interacts with the event-driven mechanism. Extending evMLP beyond classification to tasks such as detection and segmentation would further demonstrate its generality.

\section{Conclusion}
\label{sec::conclusion}

We have presented evMLP, an all-MLP architecture whose patch-independent design enables a simple event-driven local update mechanism for video processing. On ImageNet classification, evMLP achieves 73.5\% top-1 accuracy at 1.03 GMACs, outperforming similarly efficient models, and reaches 77.0\% with knowledge distillation and an extended training schedule. The event-driven mechanism reduces computational cost by 8.4\%--26.8\% on video datasets at $\tau=0$ while producing outputs identical to the dense baseline. This property follows directly from the patch-independent MLP design and distinguishes evMLP from approximate temporal-redundancy methods based on CNNs or ViTs. Increasing $\tau$ trades consistency for further savings, allowing the trade-off between efficiency and fidelity to be tuned for specific deployment scenarios.
Wall-clock measurements confirm that these savings translate into actual speedup under compute-bound conditions, and that stream-level parallelism is the effective deployment strategy for multi-core systems. The mechanism's full potential is most naturally realized on dataflow-oriented accelerators, as discussed in Section~\ref{sec::future}.

{\small
\bibliographystyle{ieee_fullname}
\bibliography{main}
}

\end{document}